\documentclass[10pt,letterpaper]{article}

\usepackage[inline]{enumitem}
\usepackage{multirow}
\usepackage[english]{babel}
\usepackage{ragged2e}
\usepackage{pifont}
\newcommand{\xmark}{\ding{53}}%
\usepackage[T1]{fontenc}    
\usepackage{hyperref}       
\usepackage{url}            
\usepackage{booktabs}       
\usepackage{amsfonts}       
\usepackage{nicefrac}       
\usepackage{microtype}      
\usepackage{multirow}
\usepackage{amsmath}
\usepackage{caption} 
\usepackage{graphicx}
\usepackage{marvosym}
\usepackage[top=0.85in,left=2.75in,footskip=0.75in]{geometry}

\usepackage{amsmath,amssymb}

\usepackage{changepage}

\usepackage[utf8x]{inputenc}

\usepackage{textcomp,marvosym}

\usepackage{cite}

\usepackage{nameref,hyperref}

\usepackage[right]{lineno}

\usepackage{microtype}
\DisableLigatures[f]{encoding = *, family = * }

\usepackage[table]{xcolor}

\usepackage{array}

\newcolumntype{+}{!{\vrule width 2pt}}

\newlength\savedwidth

\raggedright
\usepackage[aboveskip=1pt,labelfont=bf,labelsep=period,justification=raggedright,singlelinecheck=off]{caption}

\makeatletter
\renewcommand{\@biblabel}[1]{\quad#1.}
\makeatother

\usepackage{lastpage,fancyhdr}
\usepackage{epstopdf}
\usepackage{enumitem,etoolbox,calc}
\fancyheadoffset[L]{2.25in}
\fancyfootoffset[L]{2.25in}
\begin{document}
\vspace*{0.2in}

\begin{flushleft}
{\Large
\textbf\newline{Benchmarking machine learning models on multi-centre eICU critical care dataset} 
}
\newline
\\
Seyedmostafa Sheikhalishahi\textsuperscript{1,2},
Vevake Balaraman\textsuperscript{1,2},
Venet Osmani\textsuperscript{2,*}
\\
\bigskip
\textbf{1} University of Trento, Italy
\\
\textbf{2} Fondazione Bruno Kessler Research Institute, Trento, Italy
\\
\bigskip

%
%



* Corresponding author: vosmani@fbk.eu

\end{flushleft}
\section*{Abstract}
Progress of machine learning in critical care has been difficult to track, in part due to absence of public benchmarks. Other fields of research (such as computer vision and natural language processing) have established various competitions and public benchmarks. Recent availability of large clinical datasets has enabled the possibility of establishing public benchmarks. Taking advantage of this opportunity, we propose a public benchmark suite to address four areas of critical care, namely mortality prediction, estimation of length of stay, patient phenotyping and risk of decompensation. We define each task and  compare the performance of both clinical models as well as baseline and deep learning models using eICU-CRD (Collaborative Research Database) of around 73,000 patients. This is the first public benchmark on a \textit{multi-centre} critical care dataset, comparing the performance of clinical gold standard with our predictive model. We also investigate the impact of numerical variables as well as handling of categorical variables on each of the defined tasks. The source code, detailing our methods and experiments is publicly available such that anyone can replicate our results and build upon our work.


\linenumbers

\section*{Introduction}
Increasing availability of clinical data and advances in machine learning have addressed a wide range of healthcare problems, such as risk assessment and prediction in acute, chronic and critical care. Critical care is a particularly data-intensive field, since continuous monitoring of patients in Intensive Care Units (ICU) generates large streams of data that can be harnessed by machine learning algorithms. However, progress in harnessing digital health data faces several obstacles, including reproducibility of results and comparability between competing models. While, other areas of machine learning research, such as image and natural language processing have established a number of benchmarks and competitions (including ImageNet Large Scale Visual Recognition Challenge (ILSVRC) \cite{ILSVRC15} and National NLP Clinical Challenges (N2C2) \cite{stubbs2019cohort}, respectively), progress in machine learning for critical care has been difficult to measure, in part due to absence of public benchmarks. Availability of large clinical data sets, including Medical Information Mart for Intensive Care (MIMIC III) \cite{johnson2016mimic} and more recently, a multi-centre eICU-CRD (Collaborative Research Database) \cite{pollard2018eicu} are opening the possibility of establishing public benchmarks and consequently tracking the progress of machine learning models in critical care.
Availing of this opportunity, we propose a public benchmark suite to address four areas of critical care, namely mortality prediction, estimation of length of stay (LoS), patient phenotyping and risk of decompensation. We define each task and evaluate our algorithms on a multi-centre dataset of 73,718 patients (containing 4,564,844 clinical records) collected from 335 ICUs across 208 hospitals. While there has been work in this area that has focused on the single-center MIMIC III clinical dataset \cite{harutyunyan2019multitask}, our work is the first to focus on a multi-center critical care dataset, the eICU-CRD \cite{pollard2018eicu}. 
Evaluating models on a multi-center dataset typically results in the inclusion of a wider range of patient groups, larger number of patients, external validity and lower systematic bias in comparison to a single-center dataset, resulting in increased generalizability of the study \cite{bellomo2009we,youssef2008pros}. However building a predictive model on a multi-centre dataset is more challenging due to heterogeneity of the data. Nevertheless, the performance of our models (as measured by AUROC) compare favourably with the performance of the models using the single-center MIMIC-III dataset as reported in \cite{harutyunyan2019multitask}.

The main contributions of this work are as follows: i) we provide the baseline performance, (using either on clinical gold standard or Logistic/Linear Regression algorithm) and compare it against our benchmark result, achieved using a model based on bidirectional long short-term memory (BiLSTM); ii) investigate impact of categorical and numerical variables on all four benchmarking tasks; iii) evaluate entity embedding for categorical variables, versus one hot encoding; iv) show that for some tasks the number of variables can be reduced significantly without greatly impacting prediction performance; and v) report six evaluation metrics for each of the tasks, facilitating direct comparison with future results. The source code for our experiments is publicly available at \href{https://github.com/mostafaalishahi/eICU\_Benchmark\_updated}{https://github.com/mostafaalishahi/eICU\_Benchmark\_updated}, so that anyone with access to the public eICU-CRD can replicate our experiments and build upon our work.
\section*{Materials and methods}

\subsection*{Ethics statement}
This study was an analysis of a publicly-available, anonymised database with pre-existing institutional review board (IRB) approval; thus, no further approval was required.

\subsection*{eICU dataset description and cohort selection}
The eICU-CRD \cite{pollard2018eicu} 
is a multi-center intensive care unit database with high granularity data for over 200,000 admissions to ICUs monitored by eICU-CRD programs across the United States. The eICU-CRD comprises 200,859 patient unit encounters for 139,367 unique patients admitted between 2014 and 2015 to 208 hospitals located throughout the US.
We selected adult patients (age > 18) that had an ICU admission with at least 15 records, leading to 73,718 unique patients with a median age of 62.41 years (IQR, 52-75), 45.5\% female. Hospital mortality rate was 8.3\% and average LoS in hospital and in unit were 5.29 days and 3.9 days respectively (further details are provided in Table \ref{tab:gen_statistics}). Cohort selection criteria are detailed in Appendix A, Figure \ref{fig_cohort_selection}.

\begin{table}[h!]
  \centering
  \resizebox{0.8\textwidth}{!}{%
{\small
 \begin{tabular}{p{2.8cm} p{2.5cm} p{2.5cm} p{2.5cm}} 
 \hline
 & Overall & Dead at Hospital & Alive at Hospital \\ [0.5ex] 
 \hline

ICU Admissions &73,718 & 6,167 & 67,551 \\ 
Age &62.41 [52-75]&68.12 [59-80]& 61.8 [52-75]\\
Gender (F)  &33,544 (45.5) & 2,830 (45.8) & 30,714 (45.4) \\
\rule{0pt}{3ex}    
\textit{Ethnicity}\\
Caucasian &56,973 (77.2)& 4,866 (78.9) & 52,107 (77.1) \\
African American &7,982 (10.8)& 582 (9.4) & 7,400 (10.9 ) \\
Hispanic &2,937 (3.98)& 226 (3.6)& 2,711 (4) \\
Asian &1,174 (1.59) & 97   (1.5) & 1,077 (1.5) \\
Native American &413 (0.56)& 42 (0.68)& 371 (0.54) \\
Unknown &4,239 (5.7) & 354 (5.7) & 3,885 (5.7) \\
\rule{0pt}{3ex}    
\textit{Outcomes}\\
Hospital LoS* (days) &5.29 [2.53-6.84] & 3.9 [1.42-5.22] & 5.41 [2.65-6.92]\\

ICU LoS* (days) &2.32 [1.01-2.91] & 3.17 [1.19-4.43] & 2.24 [1-2.83]\\

Hospital Death &6,167 (8.36) &  6,167 (100) & -\\
ICU Death &4,575 (6.2) & 4,575 (74.1) & -\\[1ex] 
 \hline
 \end{tabular}
 }
 }
 \caption{Characteristics and mortality outcome measures. *LoS (Length of Stay). Continuous variables are presented as Median [Interquartile Range Q1–Q3]; binary or categorical variables as Count (\%)}
\label{tab:gen_statistics}
\end{table}

The final patient cohort contained 4,564,844 clinical records where we grouped these records on 1 hour window, imputed the missing values based on the mean of that window and took the last valid record of that specific window.
Out of 31 tables in the eICU-CRD (v1.0) we selected variables from the following tables: \textit{patient} (administrative information and patient demographics), \textit{lab} (Laboratory measurements collected during routine care), \textit{nurse charting} (bedside documentation) and \textit{diagnosis} based on advice from a clinician as well as consistency with other similar tasks reported in the related work section. Selected variables are shown in Table \ref{tab:features-used} and are common across all the four tasks.

\begin{table}[h!]
\centering
  \resizebox{0.4\textwidth}{!}{%
  \begin{tabular}{l l l l l l l} 
  \hline
 Variable  & Data Type \\ [0.5ex] 
 \hline
Heart rate & Numerical \\  [0.5ex] 
Mean arterial pressure & Numerical \\  [0.5ex]
Diastolic blood pressure & Numerical\\  [0.5ex]
Systolic blood pressure & Numerical\\  [0.5ex]
O2& Numerical \\  [0.5ex]
Respiratory rate& Numerical \\  [0.5ex]
Temperature & Numerical \\  [0.5ex]
Glucose& Numerical\\  [0.5ex]
FiO2& Numerical  \\  [0.5ex]
pH & Numerical  \\  [0.5ex]
Height& Numerical \\  [0.5ex]
Weight& Numerical  \\  [0.5ex]
Age & Numerical  \\  [0.5ex]
Admission diagnosis & Categorical\\  [0.5ex]
Ethnicity & Categorical\\  [0.5ex]
Gender  & Categorical \\  [0.5ex]
Glasgow Coma Score Total & Categorical\\  [0.5ex]
Glasgow Coma Score Eyes  & Categorical\\  [0.5ex]
Glasgow Coma Score Motor & Categorical\\  [0.5ex]
Glasgow Coma Score Verbal  & Categorical\\  [0.5ex]
\hline
  \end{tabular}}
   \captionsetup{justification=centering}
   \caption{Selected variables for all the four tasks}
   \label{tab:features-used}
\end{table}

\subsection*{Data Preprocessing}
The clinical variables depending on the clinical needs and the nature of the clinical variable, are measured in different intervals; for instance, the vital signs are recorded more frequently than lab measurements. In this context, to address the sparsity of data, all variables were aggregated into hourly intervals, where the last measured value was used as a candidate for that interval. In cases where the last value for each variable is not measured in the interval, the representative of that interval was computed by averaging the available measurements in the interval. Missing values that were collected hourly, like vital signs, were imputed by normal values. Categorical variables were converted into a vector to capture the semantics of each category at the model derivation phase. For all continuous variables, we utilized the recorded value in the database without any adaptation.
To address the imbalanced data in the classification tasks, we employed an over-sampling method to have an equal number of samples across different classes while providing the data as input to the models.

\subsection*{Description of tasks}
In this section, we define four different benchmark tasks, namely in-hospital mortality prediction, remaining LoS forecasting, patient phenotyping, and risk of physiologic decompensation. After applying selection criteria for each task, the resulting patient cohorts are outlined in Table \ref{tab:no_pt_rec}

\begin{table}[h!]
\centering
  \resizebox{0.8\textwidth}{!}{%
  \begin{tabular}{l c c }
  \hline
  Task& No. of patients & Clinical records\\
  \hline
  In-hospital Mortality &30,680 & 1,164,966\\
  Remaining LoS & 73,389 &3,054,314\\ 
  Phenotyping &49,299 &2,172,346\\ 
  Physiologic Decompensation &55,933 &2,800,711\\
  \hline
 \end{tabular}}
 \captionsetup{justification=centering}
   \caption{Number of patients and records in four tasks}
\label{tab:no_pt_rec}
\end{table}

\subsubsection*{Mortality prediction}
In-hospital mortality is defined as the patient's outcome at the hospital discharge. This is a binary classification task, where each data sample spans a 1-hour window. The cohort for this task was selected based on the presence of hospital discharge status in patients' record and length of stay of at least 48 hours (we focus on prediction during the first 24 and 48 hours). This selection criteria resulted in 30,680 patients containing 1,164,966 records.
\subsubsection*{Length of stay prediction}
Length of stay is one of the most important factors accounting for the overall hospital costs, as such its forecast could play an important role in healthcare management \cite{kilicc2019cost}.
Length of stay is estimated through analysis of events occurring within a fixed time-window, once every hour from the initial ICU admission. This is a regression task, where we use 20 clinical variables described in Table \ref{tab:features-used}. 
For this cohort we selected patients whose LoS was present in their records. These selection criteria resulted in 73,389 ICU stays, containing 3,054,314 records. The mean LoS was 1.86 days with standard deviation of 1.94 days, as shown in Table \ref{tab:gen_statistics}.

\subsubsection*{Phenotyping} 
Phenotyping is a classification problem where we classify whether a condition (ICD-9 code) is present in a particular ICU stay record. Since any given patient may have more than one ICD-9 code, this is defined as a multi-label classification problem. 

While our definition is focused on diagnosis using ICD codes for this task, the definition of phenotyping may encompass other domains, such as procedures \cite{shahin2019connected}\cite{mosier2020rule} for example. However, expanding the definition of phenotyping beyond standardised ICD codes would have required development of non-standardised rules, as no common standard approach for defining and validating EHR phenotyping algorithms exists \cite{denaxas2019phenotyping} \cite{denaxas2019analyzing}. Consequently, it would have been challenging to compare this work with the already published benchmarks. Furthermore, there is some concern regarding reproducibility of rule-based phenotyping as found in \cite{denaxas2019analyzing}.
Considering these issues, as well as keeping consistent with previously published benchmarks, we settled on using ICD codes as the basis for the definition of this task.
Accordingly, the dataset contains 767 unique ICD codes, which are grouped into 25 categories shown in Table \ref{tab:phen_names}. The cohort for this task, considering initial inclusion criteria as well as recorded diagnosis during the ICU stay, resulted in 49,299 patients.

\begin{table}[h!]
\centering
\resizebox{0.8\textwidth}{!}{%
\begin{tabular}{|c| p{0.8\textwidth}|} 
\hline
Type & Phenotype\\ [0.5ex] 
\hline
Acute&\begin{enumerate*}
    \item Respiratory failure; insufficiency; arrest
    \item Fluid and electrolyte disorders 
    \item Septicemia 
    \item Acute and unspecified renal failure
    \item Pneumonia
    \item Acute cerebrovascular disease  
    \item Acute myocardial infarction
    \item Gastrointestinal hemorrhage
    \item Shock
    \item Pleurisy; pneumothorax; pulmonary collapse
    \item Other lower respiratory disease
    \item Complications of surgical
    \item Other upper respiratory disease
\end{enumerate*}\\
\hline
 Chronic&\begin{enumerate*}
     \item Hypertension with complications
     \item Essential hypertension
     \item Chronic kidney disease
     \item Chronic obstructive pulmonary disease
     \item Disorders of lipid metabolism
     \item Coronary atherosclerosis and related
     \item Diabetes mellitus without complication
 \end{enumerate*}\\
 \hline
Mixed&\begin{enumerate*}
    \item Cardiac dysrhythmias
    \item Congestive heart failure; non hypertensive 
    \item Diabetes mellitus with complications
    \item Other liver diseases   
    \item Conduction disorders 
\end{enumerate*}\\
\hline
  \end{tabular}
  }
  \captionsetup{justification=centering}
  \caption{Phenotype categories}
  \label{tab:phen_names}
\end{table}

\subsubsection*{Physiologic Decompensation} \label{Decompensation-desc}
There are a number of ways to define decompensation, however in clinical setting majority of early warning systems, such as National Early Warning Score (NEWS) \cite{mcginley2012national} are based on prediction of mortality within the next time window (such as 24 hours after the assessment). Following suit and keeping consistent with previously published benchmarks \cite{harutyunyan2019multitask}, we also define decompensation as a binary classification problem, where the target label indicates whether the patient dies within the next 24 hours. The cohort for this task results in 55,933 patients (2,800,711 records), where the decompensation rate is around 6.5\% (3,664 patients).

\subsection*{Prediction algorithms}

\subsubsection*{Baselines}
We compare our model with two standard baseline approaches namely, logistic/linear regression (LR) and a 1-layer artificial neural network (ANN). The embeddings for these models are learned in the same way as for the proposed BiLSTM model as explained in the section that follows.

\subsubsection*{Deep Learning models} \label{DeepLearningModels}
\begin{figure}[h!]
    \centering
\includegraphics[width=11cm]{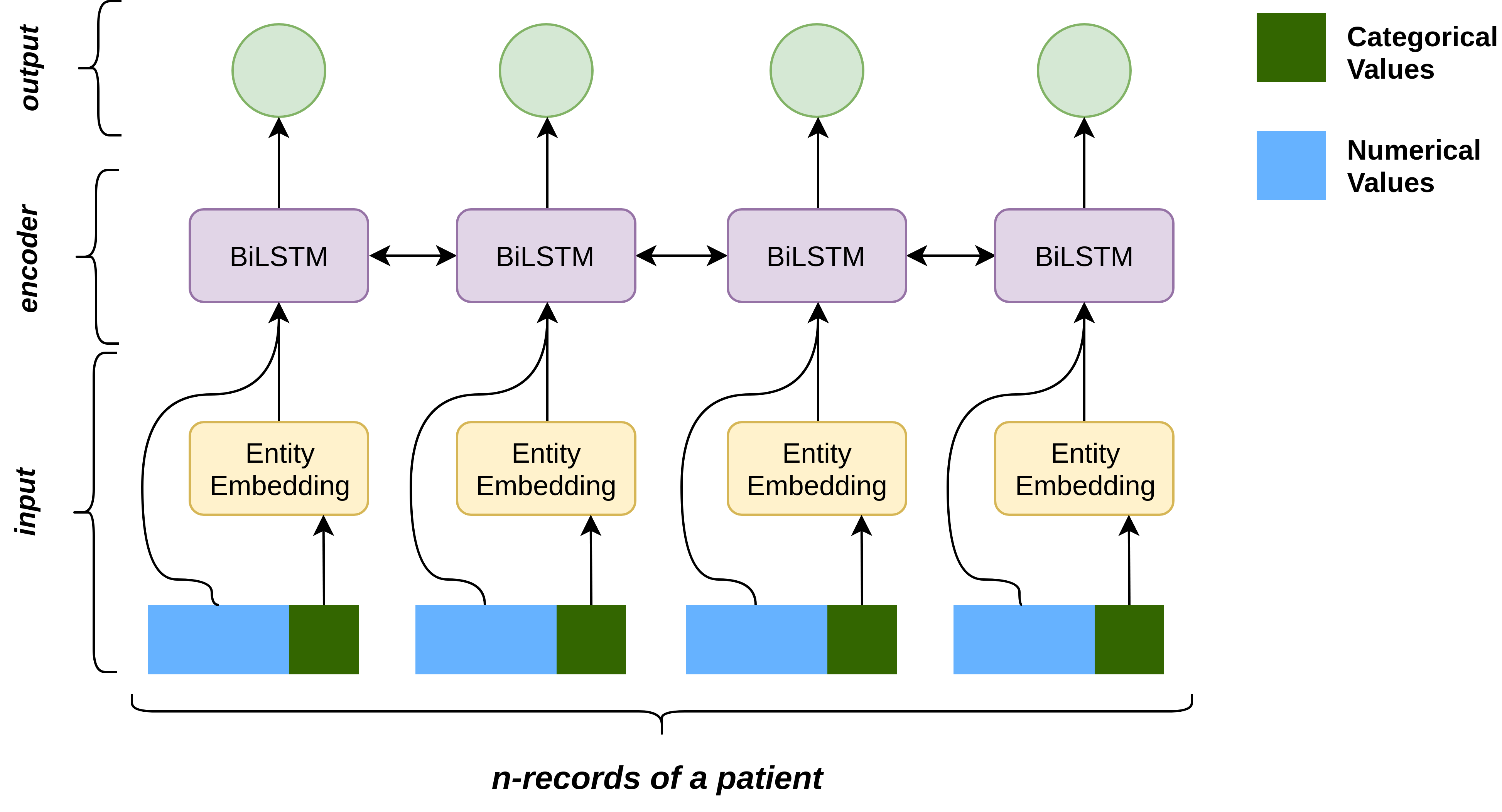}
\captionsetup{justification=centering}
  \captionof{figure}{\textbf{Model architecture}}
\label{fig_overarch}
\end{figure}

In this section, we describe the selected clinical variables, approaches to represent these variables as well as baseline and deep models used in this study. The architecture of this work consists of three modules, namely input module, encoder module and output module as shown in Fig. \ref{fig_overarch}.

\paragraph{Input representation:} We process and model both numerical and categorical variables separately, as shown in Table \ref{tab:features-used}. Categorical variables are represented using either one-hot encoding (OHE) or entity embedding (EE). OHE is the baseline approach that converts the variables into binary representation. Using this approach for our 7 categorical variables results in 429 unique records, rendering a large sparse matrix. In response, we represent each variable as an embedding and compare the performance with the OHE approach. We use entity embedding \cite{guo2016entity}, where each categorical variable in the dataset is mapped to a vector and the corresponding embedding is added to the patient's record. This entity embedding is learned by the neural network during the training phase along with other parameters.
As such, the final representation of the input at time $t$ is as follows:
\[x_t = [Num_t; U(Cat_t)]\]
where $Num_t$ is the numerical variable, $Cat_t$ is the categorical variable at time $t$ and $U$ is the embedding matrix learned by the model.

\paragraph{Encoder:}
To capture sequential dependency in our data, we use Recurrent Neural Network (RNN) that resemble a chain of repeating modules to efficiently model sequential data \cite{rumelhart1988learning}.
They take sequential data $X = (x_1, x_2,....x_n)$ as input and provide a hidden representation $H = (h_1, h_2,....h_n)$ which captures the information at every time step in the input.
Formally,
\[h_t = f(x_t + Wh_{t-1})\]
where $x_t$ is the input at time $t$, $W$ is the parameter of RNN learned during training and $f$ is a non-linear operation such as sigmoid, tanh or ReLU.

A drawback of regular RNNs is that the input sequence is fed in one direction, normally from past to future. In order to capture both past and future context, we use a Bidirectional Long Short Term Memory (BiLSTM) \cite{schuster1997bidirectional} \cite{hochreiter1997long} for our model, which processes the input in both forward and backward direction. Using a BiLSTM the model is able to capture the context of a record not only by its preceding records but also with the following records, allowing the model to produce more informed predictions.
The input at time $t$ is represented by both its forward context $\overrightarrow{h_t}$ and backward context $\overleftarrow{h_t}$ as $h_t = [\overrightarrow{h_t}; \overleftarrow{h_t}]$.
Similarly, the representation of the completed patient record is given by $h_T = [\overrightarrow{h_n}; \overleftarrow{h_1}]$.

\paragraph{Output:}
The choice of output layer is based on whether the benchmarking task is a regression or a classification task.
    
Remaining LoS prediction is a regression task, in which we predict the remaining LoS record-wise. That is, each patient record is fed to the model to predict the remaining LoS for that specific time step. This task is realized using a many to many architecture, where we assign a label to each patient record. The score for this task is obtained using:
\begin{equation}
    \widehat{y_t} = ReLU(W\cdot h_t)
\end{equation}
where $y_t$ is the remaining LoS predicted and $ReLU$ is the non-linear activation function used as the prediction of remaining LoS cannot be negative.

In-hospital mortality and decompensation are binary classification tasks. For the in-hospital mortality the many to one architecture is applied and the classifier is as follows:
\begin{equation}
    \widehat{y} = \sigma(W\cdot h_T)
\end{equation}

For the decompensation task, a many to many architecture is applied. Prediction at each-time step is treated as a binary classification and the classifier is defined as:
\begin{equation}
    \widehat{y_t} = \sigma(W\cdot h_t)
\end{equation}

Phenotyping is defined as a multi-label task with 25 binary classifiers for each phenotype, and the score for the task is obtained using:
\begin{equation}
    \widehat{y_t}^n = \sigma(W_n \cdot h_t)
\end{equation}

where $t$ is the time step and $n$ is the phenotype being predicted and $W_n$ is the model parameter.

\section*{Results}

In this section, we report benchmarking results of methods and prediction algorithms, focusing on answering the following questions: (a) How does performance of our model compare to the performance of clinical scoring systems as well as baseline algorithms (logistic/linear regression in our case); and (b) What is the impact on prediction performance when using different feature sets, such as categorical and numerical variables, solely categorical and solely numerical variables? We evaluate our model through a 5-fold cross-validation using the following evaluation metrics: for the regression task we report coefficient of determination $R^2$, and Mean Absolute Error (MAE), while for the classification tasks we report AUROC (Area Under the Receiver Operating Characteristics), AUPRC (Area Under the Precision Recall Curve), Specificity and Sensitivity (set to 90\% to facilitate direct comparison of results), Positive Predictive Value (PPV) and Negative Predictive Value (NPV); all the numerical results are presented with 95\% confidence interval (CI). 

\subsection*{Mortality prediction}
Results from this task indicate that the proposed approach of learning embeddings for categorical variables is more effective than OHE representation. This holds true for both baseline models (LR and ANN) as well as BiLSTM model, reflected in the prediction performance of each model. Furthermore, \textit{BiLSTM} model result difference is statistically significant compared to all the other approaches in predicting mortality in both the 24 hour window and the 48 hour window considering AUROC, AUPRC, and Specificity metrics as shown in Table \ref{tab:mort-risk}. It is interesting to note that using only categorical variables (reducing the number of variables from 20 to only 7) with embedding provides a better performance than using numerical variables only (AUROC 78.23\% (95\% CI, 77.08\% - 79.43\%) vs. 76.60\% (95\% CI, 76.03\% - 77.24\%) for the first 24h). These results suggest that EE of categorical features in vector space is more effective in the prediction of mortality.

\begin{table}[h!]
  \centering
  \resizebox{0.8\textwidth}{!}{
  \begin{tabular}{c| l l l l  c c c c c c} 
  \hline
 Data & Model & Num. & Cat. & Repn. & AUROC\% (95\% CI) & AUPRC\% (95\% CI)& Spec.\% (95\% CI) & Sens.\% & PPV\% (95\% CI)& NPV\%(95\% CI)\\
 \hline 
 \multirow{4}{*}{\rotatebox[origin=c]{90}{First 24 hours\hspace{1.8mm}}}
 &APACHE& \checkmark &\checkmark&Not spec.& 77.30 &  41.23& 38.74 & 86 & 57.09& 93.07\\
  & LR  & \checkmark &\checkmark&EMB&  79.95 (79.10 - 80.85)$^{\dagger}$ & 40.56 (38.84 - 42.27)$^{\dagger}$ & 46.82 (43.34 - 50.30)$^{\dagger}$ & 90 & 64.73 (60.08 - 69.37)& 90.10 (89.93 - 90.27)$^{\dagger}$\\
  & ANN  & \checkmark &\checkmark&EMB& 82.46 (81.75 - 83.24)$^{\dagger}$  &45.94 (44.77 - 47.11)$^{\dagger}$ & 50.70 (46.52 - 54.89) & 90 & \textbf{66.02 (64.01 - 68.03)} & 90.68 (90.38 - 90.98)\\
  & \textit{BiLSTM} & \checkmark & \checkmark & EMB& \textbf{83.70 (83.07 - 84.40)} & \textbf{48.47 (46.53 - 50.40)} & \textbf{53.44 (50.54 - 56.34)}& 90 & 65.91 (57.07 - 74.76) & 91.16 (90.18 - 92.13)\\
& BiLSTM & \checkmark & \checkmark  & OHE& 82.89 (82.18 - 83.65)$^{\ddagger}$ &   46.71 (44.52 - 48.90)& 50.78 (48.18 - 53.37) & 90 & 64.36 (59.12 - 69.60)& \textbf{ 91.21 (90.67 - 91.75)} \\
& BiLSTM & \xmark & \checkmark & EMB& 78.23 (77.08 - 79.43)$^{\dagger}$   &  39.89 (37.60 - 42.18)$^{\dagger}$ & 41.74 (39.58 - 43.91)$^{\dagger}$ & 90& 65.19 (53.04 - 77.34) & 90.42 (89.62 - 91.22)\\
 & BiLSTM & \checkmark &\xmark &\xmark&  76.60 (76.03 - 77.24)$^{\dagger}$  &  38.61 (36.89 - 40.34)$^{\dagger}$& 36.80 (34.40 - 39.20)$^{\dagger}$ & 90& 63.95 (60.40 - 67.50) & 90.18 (89.76 - 90.60)$^{\ddagger}$\\
  \hline
  \multirow{4}{*}{\rotatebox[origin=c]{90}{First 48 hours\hspace{2.8mm}}}
    & LR  & \checkmark &\checkmark&EMB& 82.31 (81.56 - 83.12)$^{\dagger}$ & 45.41 (44.01 - 46.80)$^{\dagger}$ & 50.96 (46.72 - 55.20)$^{\dagger}$ & 90 & \textbf{68.54 (64.91 - 72.18)} & 90.34 (90.11 - 90.57)$^{\dagger}$\\
  & ANN  & \checkmark &\checkmark&EMB& 85.27 (84.69 - 85.90)$^{\dagger}$ &52.34 (51.01 - 53.67)$^{\dagger}$ &57.31 (55.03 - 59.58) & 90 & 67.73 (64.27 - 71.19) & 91.73 (91.33 - 92.13)\\
  & \textit{BiLSTM} & \checkmark & \checkmark & EMB& \textbf{86.55 (85.65 - 87.52)} & \textbf{54.98 (53.20 - 56.77)} & \textbf{59.70 (54.85 - 64.54)}$^\dagger$& 90 & 67.16 (56.38 - 77.95) &\textbf{92.22 (90.82 - 93.63)}\\
& BiLSTM & \checkmark & \checkmark  & OHE& 85.28 (84.29 - 86.37)$^{\ddagger}$ &  52.38 (51.25 - 53.50)$^{\dagger}$& 57.22 (53.17 - 61.26) & 90 & 63.63 (56.36 - 70.90)& 92.01 (91.20 - 92.82) \\
& BiLSTM & \xmark & \checkmark & EMB & 80.33 (79.19 - 81.54)$^{\dagger}$ &  45.51 (42.88 - 48.14)& 45.51 (43.00 - 48.02)$^{\dagger}$ & 90& 63.00 (46.27 - 79.73) & 91.33 (90.12 - 92.54)\\
 & BiLSTM & \checkmark &\xmark &\xmark& 81.21 (79.75 - 82.74)$^{\dagger}$ &  	45.78 (44.07 - 47.49)$^{\dagger}$ & 47.90 (43.48 - 52.31)$^{\dagger}$ & 90& 66.30 (58.76 - 73.83) & 90.85 (90.53 - 91.17)$^{\ddagger}$\\
  \hline
  \end{tabular}
 }
  \captionsetup{justification=centering}
  \caption{In-hospital mortality prediction during first 24 and 48 hours in ICU. (\textit{Num.} and \textit{Cat.} indicate presence of numerical and categorical variables respectively. \textit{Repn.} indicates representation of categorical variables, either One Hot Encoding (OHE) or embedding (EMB) ). If the differences between \textit{BiLSTM} result and other models (LR, ANN, and BiLSTM with varying data representation) is statistically significant on a two-tailed t-test then it is indicated with $\dagger$, $\ddagger$ ($\dagger$ p < 0.05, $\ddagger$ p < 0.1). The best-performing metric values are represented in \textbf{bold} font.}
\label{tab:mort-risk}
\end{table}

\subsection*{Remaining length of stay in unit prediction}

\begin{figure}[h!]
    \centering
\includegraphics[width=8cm]{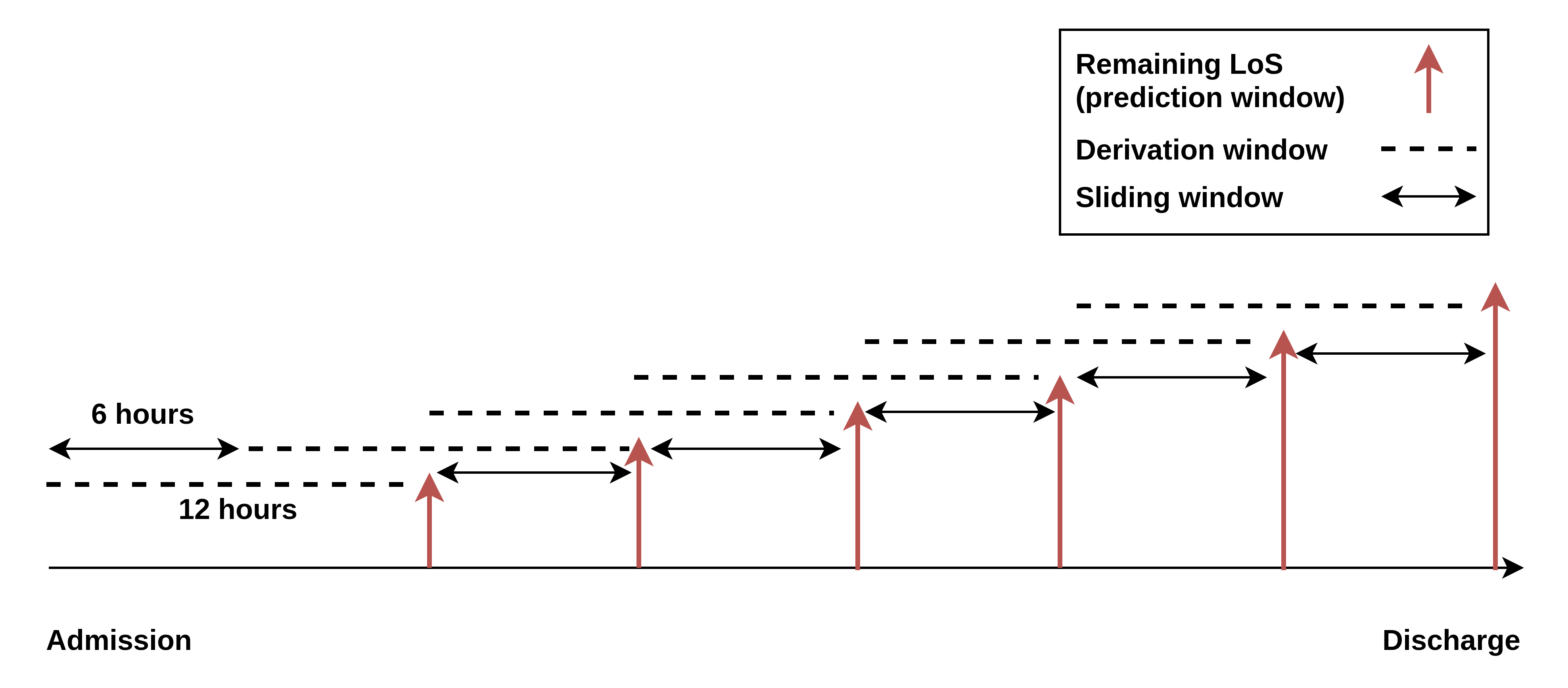}
\captionsetup{justification=centering}
  \captionof{figure}{Remaining LoS prediction schematic}
\label{fig_rolling_win_rlos}
\end{figure}

Predicting remaining LoS in the ICU with a limited number of clinical variables is a highly challenging task, as shown by the previous work \cite{harutyunyan2019multitask}; 
Our approach is designed in a way that the model requires 12 hours (derivation window) of observation data to predict the remaining LOS at the 13th hour, with a 6-hour sliding window, as shown in Figure \ref{fig_rolling_win_rlos}. By using the past 12 hours of stay in the ICU, the model creates a data representation and subsequently predicts the remaining LoS at the 13th hour up to the ICU discharge.

Results from this task indicate that
comparing \textit{BiLSTM} with ANN, we achieved incremental improvements in the $R^2$ although the result differences were not statistically significant due to the low number of time steps provided to \textit{BiLSTM}. Moreover, statistically significant differences were found comparing \textit{BiLSTM} result to LR, BiLSTM with OHE representation, BiLSTM with solely numerical, and BiLSTM with solely categorical variables in predicting remaining LoS in the ICU as shown in Table \ref{tab:los-risk-roll}. It is noteworthy that using only numerical variables provides better performance than categorical variables only ($R^2$ 0.046 (95\% CI, 0.037 - 0.056) vs. -0.003 (95\% CI, -0.042 - 0.035)). We focus the comparison of our results on $R^2$ as there is some evidence to suggest that $R^2$ metric is more informative than MAE in evaluation of regression analysis \cite{chicco2021coefficient}.

\begin{table}[h!]
  \centering
   \resizebox{0.8\textwidth}{!}{%
  \begin{tabular}{c| l l l l  c c c} 
  \hline
 Data & Model & Num. & Cat. & Repn. &  $R^2$(95\% CI) & MAE [Day](95\% CI)&\\ [0.5ex]
 \hline 
 \multirow{4}{*}{\rotatebox[origin=l]{90}{In ICU unit\hspace{1.8mm} }}
  & LR & \checkmark &\checkmark &EMB& 0.042 (0.040 - 0.044)$^{\dagger}$ & 1.301 (1.288 - 1.314) &\\
  & ANN  & \checkmark &\checkmark&EMB& 0.066 (0.056 - 0.076) &  \textbf{1.282 (1.246 - 1.319)}&\\
   & \textit{BiLSTM} & \checkmark & \checkmark &EMB&\textbf{0.075 (0.066 - 0.083)} &	1.292 (1.256 - 1.328) &  \\
   & BiLSTM & \checkmark & \checkmark&OHE& 0.054 (0.036 - 0.071)$^{\dagger}$& 1.293 (1.262 - 1.322) &   \\ 
   & BiLSTM & \xmark & \checkmark &EMB& -0.003 (-0.042 - 0.035)$^{\dagger}$	& 1.320 (1.286 - 1.354)    &\\
  & BiLSTM & \checkmark &\xmark &\xmark&  0.046 (0.037 - 0.056)$^{\dagger}$ &	1.317( 1.282 - 1.353) &\\
  \hline
  \end{tabular}}
  \captionsetup{justification=centering}
  \caption{Remaining LoS prediction in rolling window manner.
  \newline
  If the differences between \textit{BiLSTM} result and other models (LR, ANN, and BiLSTM with varying data representation) is statistically significant on a two-tailed t-test then it is indicated with $\dagger$, $\ddagger$ ($\dagger$ p < 0.05, $\ddagger$ p < 0.1). The best-performing metric values are represented in \textbf{bold} font.}
\label{tab:los-risk-roll}
\end{table}

\subsection*{Phenotyping}
For the phenotyping task, we focus on comparing performance (AUROC with 95\% CI) of the proposed model on different subset of features, namely numerical versus categorical variables.
Results from this task indicate that using both numerical and categorical features we achieved statistically significant differences compare to results from employing solely numerical and solely categorical features. Moreover, using only the categorical features, modelled as entity embeddings shows a significantly higher performance 79.40\% (95\% CI, 77.54\% - 81.26\%) compared to using only the numerical features 66.25\% (95\% CI, 63.68\% - 68.60\%) as outlined in Table \ref{tab:phen}. Clearly categorical features are more effective in representing patients' phenotype, since integrating both of the subsets does not significantly improve the result of 79.40\%(95\% CI, 77.54\% - 81.26\%) from 78.12\%(95\% CI, 76.01\% - 80.22\%).
In this task there is a wide difference between performance of the model on individual diseases, varying from 60.81\% (95\% CI, 54.86\% - 66.76\%) (diabetes mellitus without complications) to  94.34\% (95\% CI, 93.64\% - 95.04\%) (acute cerebrovascular disease). As a general trend prediction performance on acute diseases is higher (82.06\% (95\% CI, 80.58\% - 83.54\%)) than that on chronic diseases (73.03\% (95\% CI, 70.60\% - 75.46\%)). This may be due to the slow-progressing nature of chronic diseases, where recorded ICU data is relatively short and thus unable to fully capture events related to chronic diseases.

\begin{table}[h!]
  \centering
  \resizebox{\columnwidth}{!}{
  \begin{tabular}{l l l c c c} 
 \hline
 Phenotype & Prevalence & Type& \multirow{2}{*}{}&Variable\\
    \\ & & & Num \& cat  & Num. & Cat. \\ 
    &&&AUROC\% (95\% CI)&AUROC\% (95\% CI)&AUROC\% (95\% CI) \\
    [0.5ex] 
 \hline
 Respiratory failure; insufficiency; arrest  & 0.241  & acute&82.69 (81.94 - 83.44) &72.56 (71.48 - 73.64)$^\dagger$& 81.45 (81.20 - 81.70)$^\dagger$\\
 Fluid and electrolyte disorders  & 0.156  & acute & 70.92 (69.57 - 72.26)	& 60.14 (59.02 - 61.27)$^\dagger$	&71.89 (71.07 - 72.72) \\
  Septicemia  & 0.145 &  acute& 91.16 (90.56 - 91.76)&	70.56 (68.85 - 72.28)$^\dagger$& 91.14 (90.49 - 91.80)\\
 Acute and unspecified renal failure  & 0.142   & acute& 75.46 (74.70 - 76.22)&	65.20 (64.16 - 66.25)$^\dagger$&74.16 (73.09 - 75.23)$^\ddagger$\\
 Pneumonia  & 0.120  &  acute&  88.49 (87.18 - 89.79)&	69.46 (67.95 - 70.97)$^\dagger$ & 88.93 (88.19 - 89.67)\\
 Acute cerebrovascular disease  & 0.108   & acute& 94.34 (93.64 - 95.04)&73.88 (72.34 - 75.42)$^\dagger$&94.16 (93.82 - 94.50)\\
 Acute myocardial infarction  & 0.090   & acute& 91.16 (89.53 - 92.78)&	69.55 (67.75 - 71.35)$^\dagger$&	91.27 (90.14 - 92.40)\\
 Gastrointestinal hemorrhage  & 0.079   & acute & 90.43 (89.18 - 91.67)&	60.22 (58.95 - 61.48)$^\dagger$&	91.20 (90.43 - 91.96)\\
 Shock  & 0.068   & acute& 85.23 (84.62 - 85.83)&	76.61 (75.30 - 77.93)$^\dagger$ &	83.06 (82.00 - 84.11)$^\dagger$ \\
 Pleurisy; pneumothorax; pulmonary collapse & 0.039   & acute & 69.34 (67.04 - 71.63)&	59.96 (57.59 - 62.33)$^\dagger$&	70.78 (68.34 - 73.21)\\
Other lower respiratory disease  & 0.030   & acute& 80.14 (78.97 - 81.31)	& 57.24 (55.86 - 58.61)$^\dagger$&80.74 (78.98 - 82.50)\\
Complications of surgical & 0.011   & acute& 68.12 (64.25 - 71.97)&	52.27 (47.70 - 56.83)$^\dagger$&	69.03 (65.80 - 72.26)\\
Other upper respiratory disease & 0.007   & acute & 79.34 (76.36 - 82.31)&	52.94 (42.95 - 62.94)$^\dagger$&	76.23 (68.17 - 84.29)\\
 \hline
\textit{Macro-average (acute diseases)} & --- & --- & 82.06 (80.58 - 83.54)&	64.66 (62.30 - 67.02)$^\dagger$&	81.85 (80.13 - 83.56)\\
 \hline
Hypertension with complications
  & 0.019   & chronic & 84.92 (82.03 - 87.80)&79.51 (78.07 - 80.94)$^\dagger$&80.79 (78.25 - 83.32)$^\dagger$\\
Essential hypertension & 0.203   & chronic&71.00 (70.44 - 71.55)&	66.40 (65.28 - 67.51)$^\dagger$&	68.33 (67.07 - 69.59)$^\dagger$\\
Chronic kidney disease & 0.104   & chronic  & 65.83 (63.70 - 67.96)&	61.88 (60.25 - 63.50)$^\dagger$&	60.53 (58.10 - 62.96)$^\dagger$\\
Chronic obstructive pulmonary disease  & 0.093   & chronic& 76.12 (74.35 - 77.88)&	63.49 (61.78 - 65.20)$^\dagger$&	74.03 (71.16 - 76.90)\\
Disorders of lipid metabolism  & 0.054  & chronic& 72.22 (69.88 - 74.56)&	62.94 (61.81 - 64.06)$^\dagger$&	71.74 (70.47 - 73.01)\\
Coronary atherosclerosis and related  & 0.041  & chronic& 80.34 (78.97 - 81.71)&	64.46 (61.11 - 67.81)$^\dagger$&	79.55 (77.84 - 81.26)\\
Diabetes mellitus without complication  & 0.006  & chronic   & 60.81 (54.86 - 66.76)&	58.16 (50.89 - 65.43)$^\dagger$&	57.89 (51.47 - 64.31)\\
\hline
\textit{Macro-average (chronic diseases)} &---&---&73.03 (70.60 - 75.46)&	65.26 (62.74 - 67.77)$^\dagger$&	70.40 (67.76 - 73.05)\\
 \hline
Cardiac dysrhythmias & 0.165   & mixed & 74.85 (72.12 - 77.57)&	65.68 (65.11 - 66.25)$^\dagger$&71.35 (70.70 - 72.00)$^\dagger$\\
Congestive heart failure; non hypertensive  & 0.106  & mixed& 78.98 (78.16 - 79.79)&65.72 (64.94 - 66.51)$^\dagger$& 76.72 (74.13 - 79.31)$^\ddagger$\\
 Diabetes mellitus with complications  & 0.047  & mixed&92.96 (92.41 - 93.50)&	89.03 (87.08 - 90.97)$^\dagger$&	89.36 (88.05 - 90.66)$^\dagger$\\
 Other liver diseases  & 0.039   & mixed & 76.48 (73.34 - 79.62)&	68.29 (66.21 - 70.37)$^\dagger$&	76.15 (73.40 - 78.90)\\
 Conduction disorders & 0.013   & mixed  & 83.80 (80.76 - 86.84)&	67.37 (59.55 - 75.19)$^\dagger$	&82.54 (78.14 - 86.93)\\
\hline
\textit{Macro-average (mixed diseases)} & ---& --- &81.41 (79.36 - 83.46)&	71.81 (68.58 - 73.86)$^\dagger$&	79.22 (76.88 - 81.56)\\
\hline
\hline
 \textit{Macro-average (all diseases)}&---&---& 79.40 (77.54 - 81.26)&	66.25 (63.68 - 68.60)$^\dagger$&	78.12 (76.01 - 80.22)\\
 \hline
  \end{tabular}
  }
  \captionsetup{justification=centering}
  \caption{Phenotyping task on eICU-CRD (reported scores are AUROC with 95\% CI) \newline
   If the differences between \textit{BiLSTM} result and other models
  If the differences between the proposed BiLSTM model result using Num \& cat variables is statistically significant than only Num variables or Cat variables on a two-tailed t-test then it is indicated with $\dagger$, $\ddagger$ ($\dagger$ p < 0.05, $\ddagger$ p < 0.1).}
  \label{tab:phen}
\end{table}
\subsection*{Decompensation prediction}

\begin{figure}[h!]
    \centering
\includegraphics[width=8cm]{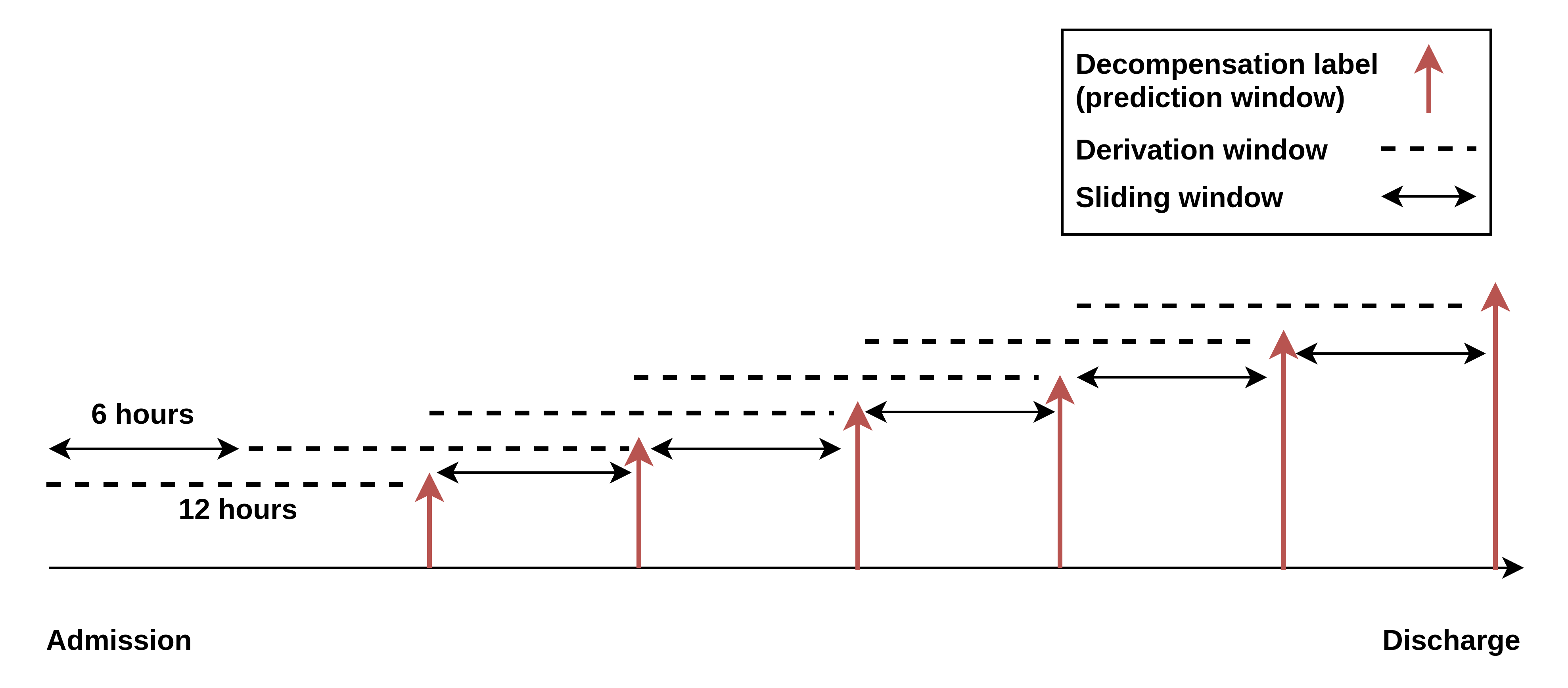}
\captionsetup{justification=centering}
  \captionof{figure}{Decompensation prediction schematic}
\label{fig_rolling_win_decomp}
\end{figure}

As mentioned in Section \nameref*{Decompensation-desc}, mortality prediction and decompensation are related, with the difference that in decompensation we predict whether the patient survives in the next 24 hours, given the previous 12 hours. Similar to the remaining LoS task, as shown in Figure \ref{fig_rolling_win_decomp} the model, using the previous 12 hours of data, predicts the decompensation state in a 6-hour rolling-window.

As demonstrated in Table \ref{tab:dec-risk-sliding} the \textit{BiLSTM} outperforms ANN although the result differences were not statistically significant due to the low number of time steps provided to \textit{BiLSTM}. Moreover, the results differences comparing \textit{BiLSTM} to LR, BiLSTM with solely categorical variables, and BiLSTM with solely numerical variables are statistically significant considering AUROC, AUPRC, and NPV metrics. Furthermore, unlike the remaining LoS task, the prediction of decompensation using the categorical variables solely outperforms employing the numerical variables solely.

\begin{table}[h!]
\centering
\resizebox{0.8\textwidth}{!}{%
\begin{tabular}{l|l c c c c c c c c c} 
\hline
Data&Model& Num. & Cat.  &Repn. &AUROC\% (95\% CI)  & AUPRC\% (95\% CI)  & Spec.\% (95\% CI) & Sens.\% & PPV\% (95\% CI)& NPV\% (95\% CI) \\
\hline
\multirow{4}{*}{\rotatebox[origin=l]{90}{In ICU unit~~}}
& LR & \checkmark & \checkmark &EMB& 78.99 (77.06 - 80.95)$^\dagger$ & 22.78 (19.88 - 25.67)$^\dagger$ & 43.92 (40.02 - 47.82)$^\dagger$ & 90.00 & 49.44 (35.70 - 63.19) & 95.08 (94.64 - 95.52) \\
& ANN & \checkmark & \checkmark &EMB& 83.57 (82.31 - 84.87) &  27.86 (24.95 - 30.78)& \textbf{54.08 (49.75 - 58.41)} & 90.00 & 48.55 (42.14 - 54.97)& 95.01 (94.71 - 95.32) \\
& \textit{BiLSTM} & \checkmark & \checkmark&EMB& \textbf{83.78 (81.86 - 85.73)} & \textbf{30.79 (28.09 - 33.49)}&  53.36 (45.92 - 60.80)& 90.00& 51.83 (43.80 - 59.86)& \textbf{95.15 (94.83 - 95.46)} \\
& BiLSTM & \xmark & \checkmark &EMB& 78.17 (75.80 - 80.56)$^\dagger$ &  20.87 (19.45 - 22.23)$^\dagger$&   42.27 (37.33 - 47.21)$^\dagger$ &90.00 & 43.55 (38.21 - 48.89)$^\ddagger$& 95.08 (94.63 - 95.52)\\
& BiLSTM & \checkmark & \xmark&\xmark& 72.96 (71.05 - 74.92)$^\dagger$&  20.00 (18.91 - 21.10)$^\dagger$& 29.05 (23.02 - 35.08)$^\dagger$ & 90.00 &\textbf{53.58 (38.43 - 68.72)} & 94.93 (94.61 - 95.25)\\
\hline
\end{tabular}
}
\captionsetup{justification=centering}
\caption{Decompensation risk prediction in eICU-CRD in a rolling-window manner\newline
If the differences between \textit{BiLSTM} result and other models (LR, ANN, and BiLSTM with varying data representation) is statistically significant
on a two-tailed t-test then it is indicated with $\dagger$, $\ddagger$ ( $\dagger$ p < 0.05, $\ddagger$ p < 0.1). The best-performing metric values are represented in \textbf{bold} font.}
\label{tab:dec-risk-sliding}
\end{table}
\section*{Discussion}

In this study we have described four standardised benchmarks in machine learning for critical care research. Our definition of benchmark tasks is consistent with previously published benchmarks to facilitate comparison with already published results. However, in this work we focus on the more recent eICU-CRD, where clinical data has been collected from 335 ICUs across 208 hospitals across the United States. Our dataset contains a larger number of patients and a wider range of patient groups, in comparison to benchmarks published using a single center dataset, which should result in lower systematic bias and increased generalisability of the study.

We provided a set of baselines for our benchmarks and show that BiLSTM model outperforms clinical gold standard as well as the baseline models. Of note is the impact of entity embedding of categorical variables in further improving the performance of our LSTM-based model. Clearly, interpretability remains a significant challenge of models based on deep neural networks, including our BiLSTM model. However, there has been significant progress in "opening the black box" \cite{zhang2018opening} as demonstrated by a recently updated review of interpretability methods \cite{molnar2019}, bringing these models one step closer to clinical practice. As our work is meant to track the progress of machine learning in critical care, interpretability is certainly an important aspect of this progress. We believe that our work will provide a solid basis to further improve critical care decision making and we provide the source code for other researchers that wish to replicate our experiments and build upon our results.
\section*{Related work}

\label{sec:related_work}
In this Section, we provide a brief review of the most relevant studies related to each of the tasks, mortality, length of stay, phenotyping, and physiologic decompensation. We briefly review the other benchmarking studies in critical care, related to our work.

\textbf{Mortality prediction}. Many clinical scoring systems have been developed for mortality prediction, including Acute Physiology and Chronic Health Evaluation (APACHE III \cite{knaus1991apache}, APACHE IV \cite{zimmerman2006acute})  and Simplified Acute Physiology Score \cite{le1993new} (SAPS II, SAPS III). Most of these scoring systems use logistic regression to identify predictive variables to establish these scoring systems. Providing an accurate prediction of mortality risk for patients admitted to ICU using the first 24/48 hours of ICU data could serve as an input to clinical decision making and reduce the healthcare costs. In this regard, recent advances in deep learning have been shown to outperform the conventional machine learning methods as well as clinical prediction techniques such as APACHE and SAPS \cite{harutyunyan2019multitask} \cite{Purushotham2018a} \cite{lipton2015learning}. Mortality prediction has been a popular application for deep learning researchers in recent years, though model architecture and problem definition vary widely. Convolutional neural network and gradient boosted tree algorithm have been used by Darabi et al. \cite{DARABI2018306}, in order to predict long-term mortality risk (30 days) on a subset of MIMIC-III dataset. Similarly, Celi et al. \cite{celi2012database} developed mortality prediction models based on a subset of MIMIC database using logistic regression, Bayesian network and artificial neural network.

\textbf{Length of stay.}
Resource allocation and identifying patients with unexpected extended ICU stays would help decision-making systems to improve the quality of care and ICU resource allocation. Therefore forecasting the length of stay (LoS) in ICU would be significantly important in order to provide high-quality care to a patient, and it would avoid extra costs for care providers. In this regard, Sotoodeh et al. \cite{sotoodeh2019improving} applied hidden markov models to predict LoS by using the first 48 hours of physiological measurements. Ma et al. \cite{ma2020length} defined LoS as a classification problem in which the objective was to create a personalized model for patients to forecast LoS. Previous studies \cite{Purushotham2018a}\cite{harutyunyan2019multitask} have shown that deep learning models obtain good results on forecasting length of stay in ICU. In this regard, Tu et al. \cite{TU1993220} applied neural network based methods on a Canadian private dataset, which includes patients with cardiac surgery. The developed model was able to detect the patient with low, intermediate, and high prolonged stay in ICU.

\textbf{Phenotyping.}
Phenotyping has been a popular task in recent years \cite{ho2014marble}\cite{zhou2014micro}, although problem definition varies widely, from focusing on ICD based diagnosis \cite{lipton2015learning} up to including clinical procedures and medications \cite{shahin2019connected} \cite{mosier2020rule}. Several works on phenotyping from clinical time series have focused on variations of tensor factorization and related models \cite{ho2014marble}\cite{zhou2014micro}\cite{kim2017discriminative}, and the most recently published studies on phenotyping are focused on deep learning methods. In this regard, Razavian et al. \cite{razavian2016multi} and Lipton et al. \cite{lipton2015learning} applied deep learning methods to predict diagnoses. While the first trained RNN LSTM and CNN for prediction of 133 diseases based on 18 laboratory tests on a private dataset including 298k patients, the latter applied an RNN LSTM on a single-center, private pediatric intensive care unit (PICU) dataset in order to classify 128 diagnoses given 13 clinical measurements.

\textbf{Physiologic decompensation.} Early detection of physiologic decompensation could be used to avoid or delay the occurrence of decompensation. Recently machine learning researchers have started to apply various machine learning methods in order to predict the decompensation incident. Recent study by Ren et al. \cite{ren2018predicting} applied gradient boosting models (GBM) to predict required intubation 3 hours ahead of time, in this work they used a cohort of 12,470 patients to predict unexpected respiratory decompensation. Differently, Xu et al. \cite{xu2018raim} proposed a deep learning model to predict the decompensation event. The proposed attention-based model was applied on MIMIC-III Waveform Database and it outperformed several machine learning and deep learning models.

\textbf{Benchmark.} Harutyunyan et al. \cite{harutyunyan2019multitask} developed a deep learning model based on RNN LSTM called multi-task RNN, in order to predict a number of clinical tasks such as mortality prediction in hospital, physiologic decompensation, phenotyping, and length of stay in ICU unit. The proposed model was applied on MIMIC-III dataset. Similarly, Purushotham et al. \cite{Purushotham2018a} have provided a single-center benchmark of several machine learning and deep learning models trained on MIMIC-III for various tasks, showing that deep learning models consistently outperformed conventional machine learning models and clinical scoring systems. One common theme across the reviewed work is that the current literature focuses on single-center databases, while we did not find any work in this area that addressed multi-centre datasets, including the associated challenges.
\section*{Acknowledgments}
We gratefully acknowledge clinical input provided by Monica Moz, Humanitas Research Hospital, Italy in both cohort selection as well as variable ranking and selection.

\section*{Appendix A}
\label{appendix_a}
\begin{figure}[h]
    \centering
\includegraphics[width=11cm]{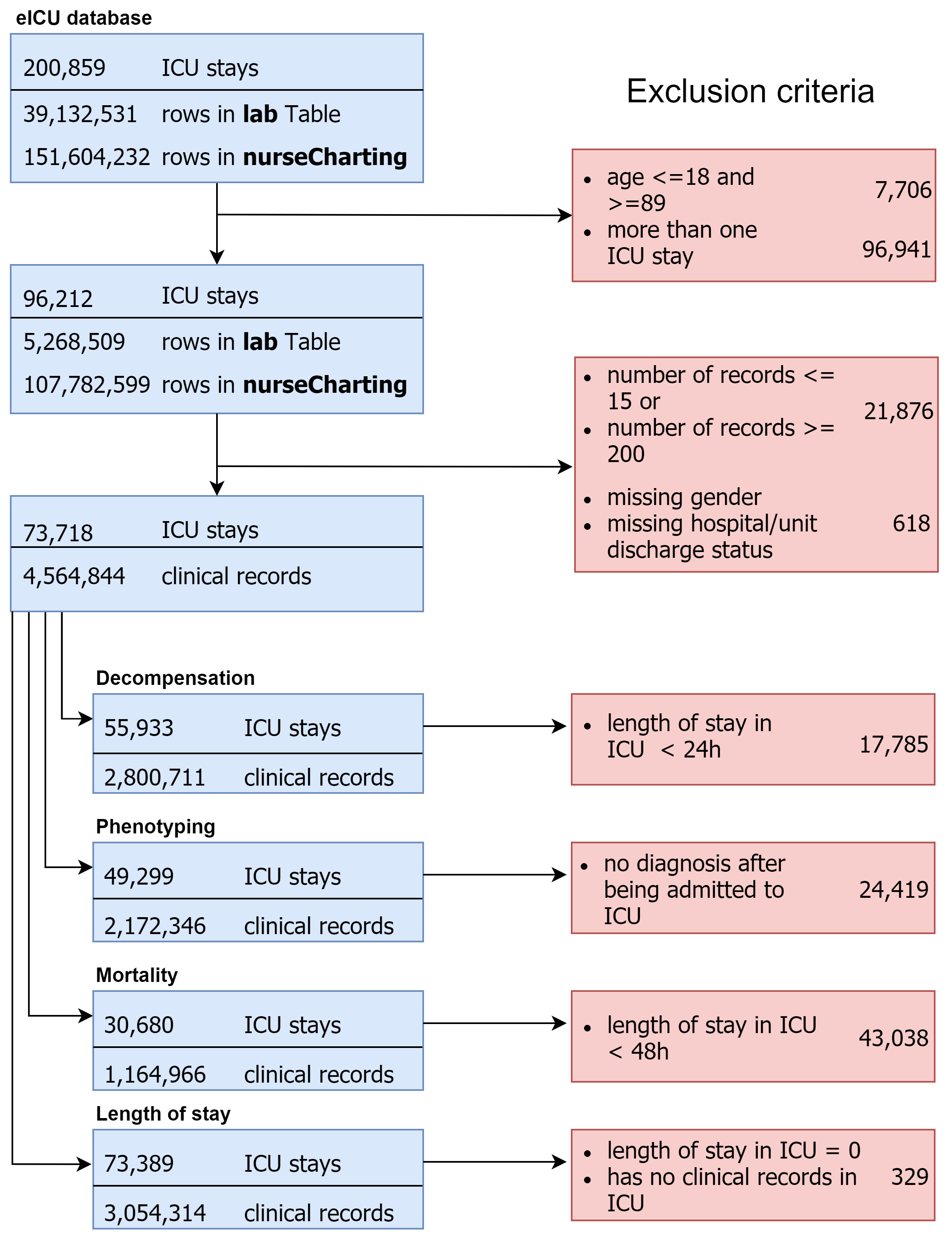}
\captionsetup{justification=centering}
  \captionof{figure}{\textbf{Cohort selection criteria}}
\label{fig_cohort_selection}
\end{figure}
\end{document}